\documentclass{article}
\usepackage{ijcai21}

\usepackage{times}
\usepackage{soul}
\usepackage{url}
\usepackage[hidelinks]{hyperref}
\usepackage[utf8]{inputenc}
\usepackage[small]{caption}
\usepackage{graphicx}
\usepackage{amsmath}
\usepackage{amsthm}
\usepackage{booktabs}
\usepackage{makecell}
\usepackage{multirow}

\makeatletter  
\newif\if@restonecol  
\makeatother

\usepackage[linesnumbered,ruled,vlined]{algorithm2e}%[ruled,vlined]{  
\usepackage{algpseudocode}  
\usepackage{amsmath}  
\title{Ensemble Federated Adversarial Training with Non-IID data}

\author{%Anonymous IJCAI submission
Shuang Luo\and
Didi Zhu\and
Zexi Li\and
Chao Wu\footnote{Corresponding author}\\
\affiliations
Zhejiang University\\
\emails
\{luoshuang, didi\_zhu, zexi.li, chao.wu\}@zju.edu.cn
}

\begin{document}

\maketitle

% 总体概览（每一个序号代表一句话）

% 摘要：
% comment
% 1、虽然联邦学习使多个分布式客户可以在保护数据隐私安全的前提要求下进行模型的协作训练，但是客户模型通常缺乏鲁棒性以至于无法抵抗对抗样本的攻击。
% 2、对抗攻击手段旨在对原始样本通过肉眼无法识别的扰动，从而在推理阶段迷惑模型使其对样本类别进行错判，这极大地影响了联邦学习中分布式客户模型的安全性。
% 3456、原来那样写就好了，也可以自己想想润色一下

% 前言：
% comment
% 第一段：
% 1、联邦学习是什么（突出保护数据安全、联合训练）：联邦学习允许各分布式客户在私有数据不泄漏的情况进行联合训练，从而在保护数据安全的前提下建立起共享的机器学习模型。
% 2、联邦学习的应用（突出安全领域，对数据安全看重，对模型安全也看中）：目前联邦学习已在各类安全领域得到了广泛的应用，如信贷风控、视觉安防、辅助诊断等，这些领域对数据私有安全性十分看重、对模型的可靠性和鲁棒性要求也非常高。
% 3、联邦学习的不足（突出模型安全的不足，就说容易受到攻击，不用说具体什么攻击）：然而虽然联邦学习保护了用户的数据安全，其忽略了模型安全层面的鲁棒性，容易受到不法分子的攻击从而导致重大安全事故的发生。

% 第二段：
% 1、对抗攻击是什么（直接介绍对抗攻击如何危害模型安全，不要说别的了）：在攻击领域最常用的一种攻击手段便是对抗攻击，对抗攻击通过xxxxx欺骗模型，直接严重地危害了模型的性能与安全性。
% 2、黑盒白盒也在这段说了它，看看怎么淡化白盒，要不直接说出于安全性的考虑，联邦学习中的模型一般不会泄漏，所以我们重点关注黑盒攻击？

% 第三段：
% 1、对抗训练是什么（介绍对抗训练帮助提高鲁棒性）
% 2、联邦学习中用对抗训练的问题（介绍直接用为什么不行，我看你说 non-IID，可以在这里展开说，这一个很重点，多想想冲突点在哪里，不要让别人觉得你是直接套娃用，而是说直接用是不可能行的，从而引出接下来的方法）

% 第四段：
% 我们提出的方法

% 第五段：
% 贡献

% 1、虽然联邦学习使多个分布式客户可以在保护数据隐私安全的前提要求下进行模型的协作训练，但是客户模型通常缺乏鲁棒性以至于无法抵抗对抗样本的攻击。
% 2、对抗攻击手段旨在对原始样本通过肉眼无法识别的扰动，从而在推理阶段迷惑模型使其对样本类别进行错判，这极大地影响了联邦学习中分布式客户模型的安全性。
% 3456、原来那样写就好了，也可以自己想想润色一下

\begin{abstract}
% While federated learning enables distributed clients to train models collaboratively under the premise of protecting data privacy and security, the clients are usually vulnerable to be attacked from adversarial samples due to the lack of robustness.
% % that are almost indistinguishable from natural data
% The adversarial attack aims to input adversarial examples to distributed clients during inference-time, resulting in mis-classification with high confidence.
% To improve the robustness of clients models, we propose the Ensemble Federated Adversarial Training (EFAT) method, a novel attack and defense federated learning algorithm. 
Despite federated learning endows distributed clients with a cooperative training mode under the premise of protecting data privacy and security, the clients are still vulnerable when encountering adversarial samples due to the lack of robustness. 
The adversarial samples can confuse and cheat the client models to achieve malicious purposes via injecting elaborate noise into normal input.  
In this paper, we introduce a novel Ensemble Federated Adversarial Training Method, termed as EFAT, that enables an efficacious and robust coupled training mechanism.
Our core idea is to enhance the diversity of adversarial examples through expanding training data with different disturbances generated from other participated clients, which helps adversarial training perform well in Non-IID settings.
Experimental results on different Non-IID situations, including feature distribution skew and label distribution skew, show that our proposed method achieves promising results compared 
with solely combining federated learning with adversarial approaches.
\end{abstract}

\section{Introduction}
%联邦学习的产生背景：设备之间如何不共享数据又能提高性能
% Mobile devices have become the primary computing resource for billions of users worldwide and billions more IoT devices are expected to come online over the next few years. These devices generate a tremendous amount of valuable data and machine learning models trained using these data have the potential to improve the intelligence of many applications. But enabling these features on mobile devices usually requires globally shared data on a server in order to train a satisfactory model. This may be impossible or undesirable from a privacy, security, regulatory or economic point of view. Therefore approaches that keep data on the device and share the model have become increasingly attractive.
% 第一段：
% 1、联邦学习是什么（突出保护数据安全、联合训练）：联邦学习允许各分布式客户在私有数据不泄漏的情况进行联合训练，从而在保护数据安全的前提下建立起共享的机器学习模型。
% 2、联邦学习的应用（突出安全领域，对数据安全看重，对模型安全也看中）：目前联邦学习已在各类安全领域得到了广泛的应用，如信贷风控、视觉安防、辅助诊断等，这些领域对数据私有安全性十分看重、对模型的可靠性和鲁棒性要求也非常高。
% 3、联邦学习的不足（突出模型安全的不足，就说容易受到攻击，不用说具体什么攻击）：然而虽然联邦学习保护了用户的数据安全，其忽略了模型安全层面的鲁棒性，容易受到不法分子的攻击从而导致重大安全事故的发生。
%联邦学习具体步骤

Federated learning is a general distributed framework that can train large-scale distributed deep learning models with a federation of participants\cite{mcmahan2017communication}. 
The central server will randomly select several clients meeting eligibility requirements in each round and broadcast the model parameters to these selected clients.
Each selected client locally computes an update based on the global model with its local dataset and then send their model parameters to the server.
The server then collects an aggregate of these updated models. It locally updates the shared model based on the aggregated update computed from the clients that participated in the current round.
As this process occurs multiple times iteratively, all clients collectively train the centralized shared model.
During this process, clients keep their private training datasets locally throughout, thereby ensuring a basic level of privacy.

Several motivating applications of federated learning in security territories have been extensively used, including human trajectory prediction \cite{10.1145/3381006},  visual inspection task \cite{10.1007/978-3-030-27272-2_5}, medical disease prediction \cite{FEKI2021107330}, etc. These domains attach high importance to data privacy and emphasize model reliability and robustness. However, although federated learning protects the security of data privacy, it is shown that clients' local deep neural network models are still vulnerable to different attacks. 

\par

% comment

 Specifically, these attack approaches can be broadly categorized into two classes:  backdoor attacks during training time and adversarial attacks during inference time. The goal of the backdoor attacks is to damage the performance of the model on targeted tasks while maintaining good performance on the main task by injecting ``poison" training data. On the other hand, adversarial attacks aim at misleading the model to misclassify the well-designed inputs called adversarial examples, which are nearly indistinguishable from raw data in human eyes but fool the trained model. This kind of attack demonstrates that these networks perform computations that are dramatically different from those in human brains. The adversary adds small perturb actions to the natural datasets that lead these systems into making incorrect predictions to achieve the goal. While the perturbations are often imperceptible or perceived as small ``noise" in the dataset, these attacks are highly effective against the deep neural network.

Up to now, a range of backdoor attacks and defense methods in federated learning settings have been introduced in previous literature, while how to defend against adversarial attack is worth considering here due to the security threat. In this paper, we are mainly concerned about improving the robustness of each node during inference time. It is still an open question whether federated learning systems can be tailored to be robust against adversarial attacks.

Adversarial attacks can be broadly classified into two types based on the knowledge of the attacked model, white-box or black-box attacks. Under white-box attacks, adversaries are necessary to have complete knowledge of the policy network, whereas black-box attacks require only access to the target model label predictions which are more applicable in many scenarios. 
The most simple yet efficient approach to perform defense is adversarial training, which injects adversarial examples into training data to fine-tuning network parameters. Nevertheless, solely adapting adversarial training to federated learning brings a range of problems. General adversarial training was developed primarily for IID data, while in federated learning, each client's data distribution is in non-IID settings. The mechanism of adversarial training in federated learning remains to be studied.

We propose the Ensemble Federated Adversarial Training (EFAT) method to improve the robustness of models against black-box attacks with non-IID training data to attack the above problems. How to resist white-box attacks is not the focus of our research because, in practice, the attacker does not know the specific parameters of the model in normal conditions. In the setting of EFAT, the central server first pre-trains the initial model on the labeled public dataset and distributes the model and the parts of the public dataset to each client. Then each client generates adversarial examples based on their public data and exchanges their adversarial examples with others during the training process. Each client performs ensemble adversarial training using their training sets and adversarial examples generated by itself and the other locally distributed public data. During this process, each client is both an attacker as well as a defender. Thus, EFAT improves the robustness of clients against adversarial attacks by enhancing the adversarial data distribution diversity.

\begin{figure*}[hbt]
    \centering
    \includegraphics[scale=1]{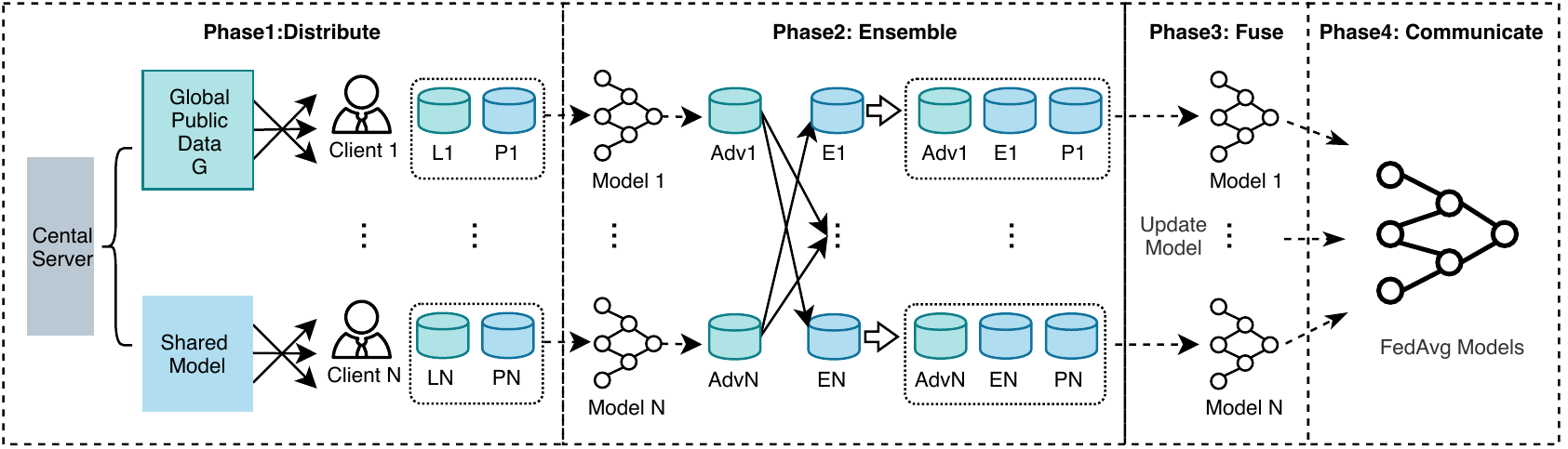}
    \caption{Illustration of the proposed EFAT framework. The EFAT method involves 4 phases: (1) \emph{Distribute}: Distributing the shared model and parts of global public dataset $L$ to all the clients. $P$ is the clients’ private data. (2) \emph{Ensemble}: Integrating adversarial samples $Adv$ generated from the local public dataset $L$ of other clients to form ensemble training data $E$. (3) \emph{Fuse}: Fusing the various data distribution including the potential knowledge of other clients by adversarial training. (4) \emph{Communicate}: Client model updates are aggregated on the central server using the FedAvg algorithm.}
    \label{fig:framework}
    \vspace{-0.3cm}
\end{figure*}

\par
\textbf{Contributions.} To the best of our knowledge, our work is the first to enhance the robustness of adversarial training in the federated learning setting by taking advantage of improving adversarial data diversity between models from distributed clients. In summary, our contributions include the following:
\begin{itemize}
    \item We explore the impact of adversarial training on the federated training paradigm and find it plays an important role. To this end, we develop a novel ensemble federated adversarial training (EFAT) methodology by incorporating adversarial examples generated by other clients' models to improve each client's robustness.
    % \item We make full use of the non i.i.d. data of clients to increase the diversity of adversarial samples，increasing the global model's robustness against black-box adversarial examples attack.
    \item Building on the above insight, we demonstrate our methodology's effectiveness and robustness against black-box attacks during inference-time on two kinds of Non-IID settings, including feature distribution skew and label distribution skew. The evaluation result shows that EFAT reaches higher adversarial accuracy on both Digit-Five and CIFAR10 than baseline.

\end{itemize}

% comment

\section{Related work}
%对抗攻击方法中再写详细一点对抗样本黑盒攻击，防御策略没必要分开 
\paragraph{Federated Learning.}
%联邦学习部分应该主要介绍联邦学习的来源和联邦学习的聚合方法进展
Federated learning has gained increasing attention in recent years due to its role in privacy protection \cite{li2020federated}. One of the most common approaches to optimizing federated learning is the Federated Averaging algorithm \cite{mcmahan2017communication}, which combines local stochastic gradient descent (SGD) on each client with a server that performs model weighted averaging with weights proportional to the size of each client’s local data.
Secure aggregation (i.e. SecAgg) \cite{bonawitz2017practical} is a tool used to ensure that the server only sees an aggregate of the client updates, not any individual client updates during FedAvg.
Several alternative aggregation schemes to address this challenge have been proposed recently due to the directly weighted averaging of model parameters that may have some adverse effects on model performance.\cite{wang2020federated}
% PFNM[64] proposes a Bayesian non-parametric approach to match clients’ weights before average, and FedMA [60] improved upon it by iterative layer-wise matching. Optimal transport[64, 74] is utilized to align individual neurons of the neural nets layer-wise before averaging. 

\paragraph{Adversarial Attack.}
Adversarial attacks refer to any alteration of the training and inference pipelines of a federated learning system designed to degrade model performance somehow. Adversarial attacks can be broadly classified into training-time attacks and inference-time attacks\cite{kairouz2019advances}.

Training-time attacks can be further classified into data poisoning \cite{bagdasaryan2020backdoor} and model update poisoning \cite{bhagoji2019analyzing,szegedy2013intriguing} based on the adversary's capability.
% Such attacks are less likely to be detected due to the adversary can only influence the data collection process at the edge of the federated learning system.
Unlike data poisoning attacks, model update poisoning attacks can directly corrupt derived quantities within the learning system.
% In model update poisoning attacks, an adversary can corrupt the updates of a client directly since the training process is done on local devices. 
% In such attacks, a backdoor can be introduced into a model even by a single-shot attack[6].
% To make matters worse, these model update poisoning attacks are even more difficult to counter when SecAgg[3] mentioned above is deployed in the aggregation of local updates.

We will concentrate on inference-time attacks and methods to defend them in this paper. Inference-time attacks generally refer to adversarial examples  \cite{szegedy2013intriguing,goodfellow2014explaining} that will be purposefully misclassified at runtime. Different from poisoning attacks, adversarial examples compromise the testing phase of machine learning. In these attacks, an adversary may attempt to circumvent a deployed model by carefully manipulating samples fed into the model. These are a perturbed version of test inputs that looks and feels the same as their original test inputs to a human, but that completely throws off the classifier \cite{goodfellow2014explaining}.
% In the image domain, adversarial examples are generally constructed by adding norm-bounded perturbations to test examples.
% In the black-box setting, models have been shown to be vulnerable to attacks based on query-access to the model [12, 13] or based on substitute models trained on similar data [8, 14, 15].
% In addition,  it is necessary to consider defense against white box evasion attacks due to any malicious client can access the model in federated learning settings.
The perturbations mentioned above can be generated by maximizing the loss function subject to a norm constraint via constrained optimization methods based gradient \cite{goodfellow2014explaining,madry2018towards}.
In the context of $l_{\infty}$ -bounded attacks, the Fast Gradient Sign Method(FGSM) \cite{goodfellow2014explaining} is one of the most popular methods using a single gradient step to perturb the inputs fed to the model.
Later, the Basic Iterative Method(BIM) \cite{kurakin2016adversarial} have been proposed, which is improved upon FGSM by applying the same step as FGSM multiple times with a small step size.
% This iterative adversarial attack was further strengthened by adding multiple random restarts.
The Projected Gradient Descent (PGD) attack, a variant of BIM, further strengthens this iterative adversarial attack by initializing examples to a random point in the ball of interest and adding multiple random restarts.
% is essentially the same as the BIM attack. The only difference is that PGD initializes the example to a random point in the ball of interest and does random restarts, while BIM initializes to the original point.
% The Projected Gradient Descent(PGD) attack[] is a variant of BIM with uniform random noise as initialization
% Projected Gradient Descent(PGD)[16] is a well-known method in the optimization literature, which applies the same step as FGSM multiple times with small step size and clip the pixel values of intermediate results after each step.
Such attacks can frequently cause naturally trained models to achieve zero accuracies on image classification benchmarks such as CIFAR10 or ImageNet \cite{carlini2017towards}, which is recognized to be one of the most potent first-order attacks.

% We firstly discuss how to defend against poisoning attacks. Existing defenses against data poisoning attacks essentially aim to sanitize the training dataset. One category of defenses [18, 19, 20, 21] detects malicious data based on their negative impact on the error rate of the learned model. Another category of defenses [22, 23, 24] leverages new loss functions, solving which detects malicious data and learns a model simultaneously. 
\par
% Federated learning is vulnerable to model evasion attacks because the deployment of the model means that adversaries have virtually no restrictions when crafting adversarial examples.
\paragraph{Adversarial Training.}
Adversarial training was first proposed by \cite{goodfellow2014explaining}, in which produced adversarial examples and injected them into original samples to strengthen a model. 
The robustness against white-box attacks achieved by adversarial training depends on the strength of the adversarial examples used.
Intuitively, adversarially trained models with FGSM or R+FGSM adversaries are only robust to single-step perturbations but remains vulnerable to more costly multi-step attacks  \cite{madry2018towards}.
To this end, adversarial training with a PGD adversary has been proposed to tackle this challenge. 
% In these settings, adversarial training is formulated as a robust optimization problem taking the form of a non-convex non-concave min-max problem, where the adversarial examples and the model weights are alternatively updated.
Since then, the PGD based adversarial training has been enhanced through various techniques, such as optimization tricks like momentum to improve the adversary, combination with other heuristic defenses like matrix estimation or logit pairing, and generalization to multiple types of adversarial attacks \cite{papernot2017practical,suciu2018does,augenstein2019generative,shafahi2019adversarial,xie2019feature}.

Despite all this, some previous work indicated that adversarially trained models might remain vulnerable to black-box attacks, where using the transferred perturbations computed on undefended models. It has been found that an adversarial network on MNIST has a slightly higher error on transferred examples than white-box examples\cite{goodfellow2014explaining}. 
Since these adversarial attacks have been observed to be transferable, adversarial training using samples generated from a single model provides robustness to other models performing the same task\cite{athalye2018obfuscated}.
To improve the robustness of black-box attacks, \cite{tramer2018ensemble} proposed an Ensembling Adversarial Training method that trained the model by injecting adversarial examples transferred from several fixed pre-trained models into the original training data. In our work, we will further strengthen the robustness of the model to defense against black-box attacks in federated learning settings.

% In addition to adversarial training,other defense methods have been proposed to make models more robust to adversarial examples.Adversarial defenses span a wide range of methods, such as differential privacy[31,32, 33].
% Morever,the researchers have also found that the defense against adversarial samples often uses obfuscation gradients, which creates a false sense of security and can actually be easily bypassed[34]. Despite the eventual defeat of other adversarial defenses, adversarial training with a PGD adversary remains empirically robust to this day.

\begin{table*}[h]
\begin{center}

\begin{tabular}{@{}llllll@{}}
\toprule
Method &
  \begin{tabular}[c]{@{}l@{}}MNIST,SVHN,\\ MNISTM,USPS\\ $\rightarrow$SYN\end{tabular} &
  \begin{tabular}[c]{@{}l@{}}MNIST,SVHN,\\ MNISTM,SYN\\ $\rightarrow$USPS\end{tabular} &
  \begin{tabular}[c]{@{}l@{}}MNIST,SVHN,\\  USPS,SYN\\ $\rightarrow$MNISTM\end{tabular} &
  \begin{tabular}[c]{@{}l@{}}MNIST,USPS,\\ MNISTM,SYN\\ $\rightarrow$SVHN\end{tabular} &
  \begin{tabular}[c]{@{}l@{}}MNIST,SVHN,\\ MNISTM,SYN\\ $\rightarrow$MNIST\end{tabular} \\ \midrule
Baseline & 61.78\% & 78.06\% & 58.61\% & 27.15\% & 90.05\% \\
EFNT     & 78.06\% & 82.50\% & \textbf{73.60\%} & 33.75\% & 98.26\% \\
EFNT+AT  & 82.30\% & 82.35\% & \textbf{73.60\%} & \textbf{48.80\%}  & 98.55\% \\
EFAT     & \textbf{85.84\%} & \textbf{83.45\%} & 71.65\% & 45.50\%  & \textbf{98.65\%} \\ \bottomrule
\end{tabular}
\caption{Performance of clients trained with Digit-Five against black-box PGD adversaries. We assign four of them to conduct different adversarial training methods while the rest are regarded as adversaries to perform black-box PGD-40 attacks.
Our proposed methods EFNT, EFNT+AT, and EFAT outperform the baseline among these five different dataset extraction schemes. In three-fifths of the data set extraction situations, EFAT reached the highest robust accuracy.}
\label{table:DigitFive}
\end{center}
\vspace{-0.5cm}
\end{table*}

\section{Methodology}
We consider the notations and definitions of federated learning as defined in \cite{mcmahan2017communication}. 
To be specific, there are  $K$ clients connected to a central server in federated learning.  At each round $t$, the server randomly selects $ N = \lfloor C \times K \rfloor$ clients for some $0 < C < 1$.We assume that for every $ 1 \leq i \leq N $ the $i_{th}$ node has access to private training samples in $P_i=\{ (x_i,y_i) \in R^d \times R \}$.
\subsection{Intuitive Federated Adversarial Training}
In this paper, we mainly study how to adapt adversarial training to federated learning.  We first introduce the intuitive federated adversarial training method combining federated learning with adversarial training directly.

\paragraph{Local PGD Attack}
For each selected client node in the federated learning model, we perform a PGD attack locally to generate their own basic adversarial examples.

The PGD attack first performs a gradient ascent step in the loss function w.r.t. the image pixel values. For the $i_{th}$ client, PGD attack performs multi-steps update on the original sample $x_i$ along the direction of the gradient of a loss function.

We use local private dataset $ P_i $ as input of the client model $M^i$. In each iteration, PGD adversarial examples $P_i^{adv}=\{ x_i^{adv},y_i \}$ follows the update rule:
\begin{equation}
    x^{adv}_{t+1}=\Pi_{clip}(x^{adv}_t+\alpha sign(\nabla_{x} J(x_t^{adv},y) ))
\end{equation}
where $ \alpha $ controls the maximum $L_{\infty}$ perturbation of the adversarial examples, and the clip function forces $x$ to reside in a certain range.

\paragraph{Local Adversarial Training}
Follow the work of \cite{kurakin2016adversarial}, we first group examples into batches containing both normal and adversarial examples before taking each training step. We use a loss function that allows independent control of the number of adversarial examples in each batch:
\begin{equation}
    Loss=\sum_{x_i \in P_i} L(x_i|y_i)+\sum_{x^{adv}_i \in P_i^{adv}}L(x^{adv}_i|y_i)
\end{equation}
where $L((x|y)$ is a loss on a single example $x$ with true class $y$.

After the loss function is determined, we perform adversarial training on the $i^{th}$  local client, which can be formulated as a robust optimization problem \cite{madry2018towards,tramer2018ensemble}.
\begin{equation}
    \min_{\theta^i} \max_{D(x^i,x^i_{adv})<\alpha} Loss
\end{equation}
 The inner maximization problem synthesizes the adversarial counterparts of clean examples, while the outer minimization problem finds the optimal model parameters over perturbed training examples. 
\paragraph{Federated Averaging}
After each selected client performs local adversarial training, these client model updates are sent to the central server. The central server then aggregates these local model parameters using the FedAvg \cite{mcmahan2017communication} algorithm. The global averaging step in time step t can be written as follows:
\begin{equation}
    \theta_{t} = \sum_{i=1}^{N}\frac{1}{N}\theta^i_{t} 
\end{equation}

 Thus, each local robust client model is trained individually. Obviously, this training paradigm only considers the client-specific loss, which leads to the federated model being still vulnerable against adversarial examples generated with other models.
 
\subsection{Ensemble Federated Adversarial Training}

\par
In our proposed ensemble federated adversarial training (EFAT) method, to tackle the challenge mentioned above, we take advantage of extensive and diverse knowledge from other clients to improve the robustness of previously limited models to small populations (Algorithm \ref{alg::conjugateGradient}).
% In this section, we will introduce the core idea of the proposed ensemble federated adversarial training(EFAT) method and describe how to implement this approach. 
The EFAT method can be summarized 4 phases as follows(see Figure \ref{fig:framework}): (1) \emph{Distribute}: Distributing the shared model and parts of global public dataset to all the clients. (2) \emph{Ensemble}: Integrating adversarial samples generated from the local public dataset of other clients. (3) \emph{Fuse}: Fusing the various data distribution including the potential knowledge of other clients by adversarial training. (4) \emph{Communicate}: Client model updates are aggregated on the central server using the FedAvg algorithm.

\begin{algorithm}  
  \caption{Illustration of EFAT on $K$ homogeneous clients (indexed by k) for T rounds, $n_k$ denotes the number of data points per client and $C$ the fraction of clients participating in each round. The server model is initialized as $x_0$. $G$ is the global public dataset, $L$ is the local public dataset, and $P$ is the client private data.}  
  \label{alg::conjugateGradient}
  % \KwIn{$r_i$, $Backgrd(T_i)$=${T_1,T_2,\ldots ,T_n}$ and similarity threshold $\theta_r$}  
  % \KwOut{$con(r_i)$}  
  \textbf{Server executes:}
  
  \For{each communication round $t = 1,...,T$}  
  {  
    $S_t$ $\gets$ random subset ($C$ fraction) of $K$ clients  

    distribute part of G to $K$ clients \{$L_1, L_2, ..., L_k$\}

    \For{each client $k$ $\in$ $S_t$ in parallel}  
    {  
      generate adversarial examples $L_k^{adv}$ by Eq.(1)
      
      exchanges with other clients $E_k^{adv}$ = \{$L_1^{adv},..., L_{k-1}^{adv}$\}

      ${x_{t+1}^k}$ $\gets$ Client-localSGDupdate($k$, $x_t$) using ensemble \{$P_k$, $L_k^{adv}$, $E_k^{adv}$\}
    }  
    $x_{t+1} \leftarrow \sum_{k=1}^{K} \frac{n_{k}}{n} x_{t+1}^{k}$\;  
  }  
  return $x_{t+1}$\;

  % \textbf{ClientUpdate($k$,$x_t$):}

  % \For{each local epoch $i$ from $1$ to $E$}  
  % {  
  %   Do one training step of network $x_t$ using ensemble \{$P_i$, $L_i^{adv}$, $E_i^{adv}$\}
 
  % }  
  % return $x_{t+1}$\; 
  
\end{algorithm}  

% \begin{algorithm}[h]  
%   \caption{Illustration of EFAT on $K$ homogeneous clients (indexed by k) for T rounds, $n_k$ denotes the number of data points per client and $C$ the fraction of clients participating in each round. The server model is initialized as $x_0$. $G$ is the public dataset, and $L$ is the local public dataset.}  
%   \label{alg::conjugateGradient}  
%   \begin{algorithmic}[algorithm1] 
%   % \Procedure{Server executes:}{}  
%   \State \textbf{Server executes:}
%     \State for each communication round $t = 1,...,T$ do 
%     % \Repeat
%     \State $S_t$ $\gets$ random subset ($C$ fraction) of $K$ clients
%     \State for each client $k$ $\in$ $S_t$ in parallel do
%     \State ${x_{t+1}^k}$ $\gets$ ClientUpdate($k$, $x_t$)

% \State \textbf{ClientUpdate($k$,$x_t$):}
%       \State for each local epoch i from 1 to E do
%       \State generate adversarial examples $L_i^{adv}$ by $L_i$
%       \State Do one training step of network $x_t$ using ensemble \{$P_i$, $L_i^{adv}$, $E_i^{adv}$\}
%     % \Until{($f(x_k)>f(x_{k-1})$)}  
%   % \EndProcedure
%   \end{algorithmic}  
% \end{algorithm}  

\paragraph{Distribute}
In this stage, the pre-training shared model and the part of public dataset are distributed to all clients. Therefore, the training dataset in each client is partitioned into two parts:(1) the client part $P$ and (2) the public part $G$. $P$ is partitioned into participated clients taking feature distribution skew and label distribution skew into consideration. $G$ is the globally public dataset that consists of a uniform distribution over features or labels. A random $\alpha$ proportion of the global public dataset $G$ is distributed to each client. It can be concluded that the data owned by each client consist of the private data $P$ and a random $\alpha$ proportion of $G$. We denote the subset of $P$ and a random $\alpha$ proportion of $G$ by \textbf{\emph{privated data $P_i$}} and \textbf{\emph{local public data $L_i$} }in the $i_{th}$ client respectively. 

\paragraph{Ensemble}
\par
We use the local public data $L_i$ to generate the corresponding adversarial examples denoted by \emph{local adversarial public data $L_i^{adv}$} .Then we can get \emph{  ensemble adversarial public data $\{L_1^{adv}, L_2^{adv},...L_{i-1}^{adv},L_{i+1}^{adv},... L_N^{adv} \}$  } by ensembling the local adversarial public data generated by other clients except the current $i_{th}$ client. We denote ensemble adversarial public data 
as $E_i^{adv}$.
In this stage, we can conclude that the data composition of the $i_{th}$ client consists of three parts: (1) private data, (2) local adversarial public data, (3) ensemble adversarial public data. 
%  In this stage,we augments training data with adversarial inputs generated from other client models to train some client model.
% In our ensemble adversarial training algorithm, we decouple the adversarial examples from the current model, while simultaneously drawing an explicit connection with robustness to black-box adversaries.
% Correspondingly, the inputs of the client model $M^i$ are composed of $S^i$ (the original local dataset) and $A^1 ,..., A^N $(the adversarial examples over the current model and other client models). 
\begin{equation}
    Data_{i}=\{P_i,L_i^{adv},E_i^{adv}\}
\end{equation}

\paragraph{Fuse}
In this stage, we perform adversarial training on both local adversarial public data, ensemble adversarial public data and private data.
It should be noted that the loss function will change correspondingly compared with the previous equation (2) due to different data distributions.
\begin{equation}
\begin{split}
    Loss=\sum_{x_i \in P_i} L(x_i|y_i)+\sum_{x^{adv}_i \in L_i^{adv}}L(x^{adv}_i|y_i) \\ +\sum_{x^{adv}_i \in E_i^{adv}}L(x^{adv}_i|y_i)
\end{split}
\end{equation}
Intuitively, as adversarial examples transfer between models, perturbations crafted on other clients are good approximations for the maximization problem. Moreover, the learned model can not influence the “strength” of these adversarial examples. As a result, minimizing the training loss implies increased robustness to black-box attacks from other models.
Through this ensemble method, we can take advantage of more extensive and more diverse datasets to improve the robustness of previously limited models to small populations.

% \begin{table*}
% \begin{center}
% \begin{tabular}{llllll}

% Method &
%   \begin{tabular}[c]{@{}l@{}}MNIST,SVHN,\\ MNISTM,USPS\\ $\rightarrow$ SYN\end{tabular} &
%   \begin{tabular}[c]{@{}l@{}}MNIST,SVHN,\\ MNISTM,SYN\\ $\rightarrow$USPS\end{tabular} &
%   \begin{tabular}[c]{@{}l@{}}MNIST,SVHN,\\  USPS,SYN\\ $\rightarrow$MNISTM\end{tabular} &
%   \begin{tabular}[c]{@{}l@{}}MNIST,USPS,\\ MNISTM,SYN\\ $\rightarrow$SVHN\end{tabular} &
%   \begin{tabular}[c]{@{}l@{}}MNIST,SVHN,\\ MNISTM,SYN\\ $\rightarrow$MNIST\end{tabular} \\
% Baseline & 61.78 & 78.06 & 58.61 & 27.15 & 90.05 \\
% EFNT     & 78.06 & 82.50 & 73.60 & 33.75 & 98.26 \\
% EFNT+AT  & 82.30 & 82.35 & 73.60 & 48.8  & 98.55 \\
% EFAT     & 85.84 & 83.45 & 71.65 & 45.5  & 98.65
% \end{tabular}
% \end{center}
% \end{table*}

\paragraph{Communicate}
After each selected client performs local update based on the ensemble adversarial training, these model updates are sent to the server. Then the central server aggregates these models by averaging to obtain the new global model.

In summary, we first assign the shared model and part of the public dataset to participated clients. Then we ensemble adversarial examples generated from local public data distributed on multiple clients. Next,we perform adversarial training based on these perturbations. The last step is to average local model updates using FedAvg algorithm. This training procedure given above is repeated until a satisfactory degree of convergence has been achieved.

\section{Experiment}
This section demonstrates the robustness against black-box attacks of our proposed algorithm on two kinds of highly Non-IID datasets, including feature distribution skew and label distribution skew. 
All experiments were done using a V100 GPU cluster, and the federated system was simulated on a single machine (as the communication efficiency is not the main focus of this paper). Experiment results show that our models significantly improve robustness and accuracy against black-box attacks, which provides strong support for our central hypothesis.

\subsection{Experiment Setup}

\begin{table*}[]

  \centering
\begin{tabular}{@{}llllllllll@{}}
\toprule
\multicolumn{1}{c}{\multirow{2}{*}{non-i.i.d.-ness}} & \multicolumn{3}{c}{IID}              & \multicolumn{6}{c}{Non-IID}                                                 \\ \cmidrule(l){2-10} 
\multicolumn{1}{c}{}                            & \multicolumn{3}{c}{$\gamma$=100}            & \multicolumn{3}{c}{$\gamma$=1}              & \multicolumn{3}{c}{$\gamma$=0.01}           \\ \midrule
\multicolumn{1}{l}{Method}   &
  \multicolumn{1}{c}{clean} &
  \multicolumn{1}{c}{PGD-10} &
  \multicolumn{1}{c}{PGD-20} &
  \multicolumn{1}{c}{clean} &
  \multicolumn{1}{c}{PGD-10} &
  \multicolumn{1}{c}{PGD-20} &
  \multicolumn{1}{c}{clean} &
  \multicolumn{1}{c}{PGD-10} &
  \multicolumn{1}{c}{PGD-20} \\ \midrule
\multicolumn{1}{l}{Baseline}                                         & 72.21\%          & 62.34\% & 64.17\% & 72.21\%          & 63.02\% & 63.62\% & 72.46\%          & 65.05\% & 64.62\% \\
\multicolumn{1}{l}{EFNT}                                              & \textbf{80.45\%} & 43.91\% & 41.02\% & \textbf{79.03\%} & 43.23\% & 43.99\% & \textbf{81.19\%} & 47.82\% & 42.79\% \\
\multicolumn{1}{l}{EFNT+AT}                                           & 78.42\%          & 57.45\% & 48.35\% & 78.81\%          & 58.36\% & 46.41\% & 81.15\%          & 56.24\% & 49.41\% \\
\multicolumn{1}{l}{EFAT}    &
  72.02\% &
  \textbf{70.83\%} &
  \textbf{67.25\%} &
  72.64\% &
  \textbf{70.28\%} &
  \textbf{68.64\%} &
  74.66\% &
  \textbf{71.48\%} &
  \textbf{67.46\%} \\ \bottomrule
\end{tabular}
\caption{Accuracies of CIFAR10 under black-box attacks in IID and non-IID settings w.r.t. $\alpha$ = 5\%. Our proposed defense method can significantly improve the robust test accuracy of deep models on clients in both IID and non-IID settings with $\alpha$ are set to 5\%.  EFAT is able to achieve robust test accuracy as high as 71.48\%, increasing 5\%-8\%,24\%-27\% and 13\%-28\% compared to the baseline method, EFNT and EFNT+AT method respectively. }
\label{table:beta5}
\vspace{-0.1cm}
\end{table*}

\begin{table*}[]

  \centering
\begin{tabular}{@{}llllllllll@{}}
\toprule
\multicolumn{1}{c}{\multirow{2}{*}{non-i.i.d.-ness}} & \multicolumn{3}{c}{IID}                          & \multicolumn{6}{c}{Non-IID}                               \\ \cmidrule(l){2-10} 
\multicolumn{1}{c}{}                            & \multicolumn{3}{c}{$\gamma$=100}                       & \multicolumn{3}{c}{$\gamma$=1}    & \multicolumn{3}{c}{$\gamma$=0.01}  \\ \midrule
\multicolumn{1}{l}{Method} &
  \multicolumn{1}{c}{clean} &
  \multicolumn{1}{c}{PGD-10} &
  \multicolumn{1}{c}{PGD-20} &
  \multicolumn{1}{c}{clean} &
  \multicolumn{1}{c}{PGD-10} &
  \multicolumn{1}{c}{PGD-20} &
  \multicolumn{1}{c}{clean} &
  \multicolumn{1}{c}{PGD-10} &
  \multicolumn{1}{c}{PGD-20} \\ \midrule
\multicolumn{1}{l}{Baseline}                   & 72.91\% & 62.46\% & \multicolumn{1}{l}{61.79\%} & 72.95\% & 61.83\% & 60.41\% & 72.93\% & 61.29\% & 59.80\% \\
\multicolumn{1}{l}{EFNT}                       & \textbf{82.32\%} & 43.51\% & \multicolumn{1}{l}{43.49\%} & \textbf{80.23\%} & 50.17\% & 44.84\% & \textbf{82.89\%} & 45.36\% & 42.20\% \\
\multicolumn{1}{l}{EFNT+AT}                    & 81.45\% & 65.26\% & \multicolumn{1}{l}{62.45\%} & 79.54\% & 64.65\% & 62.12\% & 81.57\% & 65.04\% & 62.12\% \\
\multicolumn{1}{l}{EFAT}  &
  73.39\% &
  \textbf{70.47\%} &
  \textbf{68.35\%} &
  73.45\% &
  \textbf{70.98\%} &
  \textbf{68.25\%} &
  75.57\% &
  \textbf{71.66\%} &
  \textbf{68.43\%} \\ \bottomrule
\end{tabular}
\caption{Accuracies of CIFAR10 under black-box attacks in IID and non-IID settings w.r.t. $\alpha$ = 10\%.  A different trend can be observed where the robust test accuracy of EFNT+AT increases 10\%-18\% when defending adversarial examples generated by PGD-10 and PGD-20. }
\label{table:beta10}
\vspace{-0.4cm}
\end{table*}

\paragraph{Datasets}
% In federated learning practice, data in each node are usually inherently non-iid, which means the nodes may hold feature variants or label shift data. To simulate the practical scenarios, we use Digit-Five datasets, which is a collection of five benchmarks for digit recognition, namely MNIST (LeCun et al., 1998), Synthetic Digits (Ganin & Lempitsky, 2015), MNIST-M (Ganin & Lempitsky, 2015), SVHN, and USPS.
We use Digit-Five datasets as feature distribution skew datasets, which is a collection of five benchmarks for digit recognition, namely MNIST \cite{lecun1998gradient}, Synthetic Digits \cite{ganin2015unsupervised}, MNIST-M \cite{ganin2015unsupervised}, SVHN, and USPS. It was constructed for domain adaptation research by \cite{peng2020federated}.

We construct a label distributed skew dataset based on CIFAR10 by using the Dirichlet distribution \cite{lin2020ensemble}. The value of $\gamma$ controls the degree of non-i.i.d.-ness. When $\gamma$ tends to 0, the clients are more likely to hold examples from only two classes (if the number of clients is set to 5). Besides, $\gamma = 100$ mimics identical local data distributions. We conduct comparative experiments using three different $\gamma$ values, respectively 100, 1, 0.01.

\paragraph{Training Strategy}
For Digit-Five, we take turns selecting four datasets as different participated clients. Then we assign $10\%$ of $G$ to each client as local public dataset. The four participated clients perform ensemble federated adversarial training while the unselected dataset trains by itself and then generates adversarial examples as black-box attacks to test our model's performance.

For CIFAR10, the training dataset consists of 50000 images in 10 classes, with 5000 images per class. The training dataset are distributed to 5 clients using a Dirichlet distribution mentioned above. Each client can get about 10000 (50000/5) images as private data. We set the random distributed fraction $\alpha$ as $10\%$. Then we assign $10\%$ of $G$ to each client as local public data. 

In the experimental setting, the clients train locally for five rounds and then exchanges local public adversarial datasets once.

\paragraph{Compared Methods}
 To illustrate the necessity and effectiveness of our training method in detail, we introduce two different simplified versions of EFAT called EFNT and EFNT+AT.
 
% In our naming rules，“NT” refers to normal training and “AT” refers to adversarial training.
%  In addition, each client node consists of three parts:private data,local shared data and ensemble shared data.
%  \begin{itemize}
%      \item EFAT:private data NT, local shared data AT, ensemble shared data NT.
%      \item EFNT+AT:private data NT, local shared data NT, ensemble shared data NT.
%      \item EFNT:private data NT, local shared data NT, ensemble shared data NT.
%  \end{itemize}
In our EFAT method, we perform adversarial training on both local adversarial public data and ensemble adversarial public data.
EFNT+AT refers to performing normal training instead of adversarial training on ensemble public data.
EFNT refers to performing normal training instead of adversarial training to ensemble public data and local public data.

Besides, we adopt the intuitive federated adversarial learning mentioned in Section 3.1 as the baseline method. For a fair comparison, we extract the same amount of private data as the sum of local public data and ensemble public data to generate adversarial examples for adversarial training.

\paragraph{Networks and Parameters}
In the experiments, we simulate a federated learning scenario with $n = 4$ nodes where each node uses ResNet18 with the same architectures. We choose to take gradient steps in the $L_{\infty}$ norm, i.e., adding the sign of the gradient, since this makes the choice of the step size simpler.

% We perform PGD (Kurakin et al., 2017) with 10 and 20 steps towards the aggregated server model.  All PGD attacks have random start, i.e, the uniformly random perturbation of [-$\epsilon_{test}$, $\epsilon_{test}$] added to the clean test data before PGD perturbations. 

For Digit-Five, we set perturbation $\epsilon = 0.3$, perturbation step size $\eta_1 = 0.01$,number of iterations $K = 40$, learning rate $\eta_2 = 0.01$,batch size $m = 128$, and run 100 epochs on the training dataset. To evaluate robust errors, we apply PGD (black-box) attack with 20 and 40 iterations and 0.01 step size. 
For CIFAR10, following \cite{tramer2018ensemble}, the maximum perturbation allowed is 16/255 for both defense and attack models. We set perturbation $\epsilon_{train}$ = 16/255, step size $\alpha$ = 0.003, number of iterations K = 20, batch size m = 128, and run 100 epochs on the training dataset.The adversarial test data are bounded by  $L_{\infty}$ perturbations with $\epsilon_{test}$ = 16/255 and 8/255  which are generated by PGD-10 and PGD-20.

All PGD attacks have a random start, i.e., the uniformly random perturbation of [-$\epsilon_{test}$, $\epsilon_{test}$] added to the clean test data before PGD perturbations.

\subsection{Result Analysis}
\paragraph{Digit-Five}
In the following experiment, we performed our proposed EFAT, EFNT, EFNT+AT, and baseline with Digit-Five datasets on both the ``clean" examples $x$ and adversarial examples $x_{adv}$. Table \ref{table:DigitFive} illustrates each client's average accuracy against PGD-40 black-box attacks.
For example, the first column means that we select MNIST, SVHN, MNIST-M, and SVHN and distribute them to four clients to perform different training methods. Simultaneously, SYN trains by itself and then generates adversarial examples sending to the first four clients as a black-box attack.
A first observation is that compared with the models only trained locally with their own adversarial examples(baseline), EFAT, EFNT, and EFNT+AT trained with exchange public data reach higher accuracy against attacks. It is because in the setting of feature distribution skew, expanding training data from different clients' models increases the diversity of training data distribution, which helps improve the robustness of models.

We also noted that EFAT outperforms EFNT and EFNT+AT, which indicates adversarial examples generated for one model could stay adversarial for other models.
Therefore, it is helpful when conducting adversarial training using adversarial examples generated by other clients' public data.

\paragraph{CIFAR10}

For CIFAR10, we compare our EFAT algorithm with the baseline method, EFNT, and EFNT+AT, with three different $\gamma$ values, respectively $100, 1, 0.01$. 
Table \ref{table:beta5} shows the performance of EFAT and the other three methods w.r.t. standard test accuracy and adversarially robust test accuracy of the clients on CIFAR10. It should be noted that the accuracy here refers to the average accuracy of all participated client models.
% Table x shows the performance on Digit-Five.
We obtain standard test accuracy for clean test data and robust test accuracy for adversarial test data generated by PGD-10 and PGD-20.

From Table \ref{table:beta5} we can observe that our proposed defense method can significantly improve the robust test accuracy of deep models on clients in both IID and non-IID settings with $\beta$ are set to 5\%.
Our EFAT method is able to achieve robust test accuracy as high as 71.48\%, increasing 5\%-8\%,24\%-27\% and 13\%-28\% compared to the baseline method, EFNT and EFNT+AT method respectively. This gap significantly widens as "non-i.i.d.-ness" (Specifically refers to perturbation bound $\epsilon_{test}$) increases. Larger “non-i.i.d.-ness”  will allow the generated adversarial data to deviate more from natural data. In EFNT and EFNT+AT methods, robust test accuracies are significantly hurt with larger $\epsilon_{test}$. 

In addition, client models trained with EFNT method achieve the highest clean test accuracy, followed by EFNT+AT method, while the baseline method and our EFAT method do not perform well. 
It is forgivable that adversarial training provides the most security of adversarial attacks while losing only a small amount of accuracy when we mainly focus security against adversarial examples.

Besides, we compare our EFAT method and other methods with different values of $\alpha$. We set $\alpha$ = 10\% in Table \ref{table:beta10}. We observe a different trend where the robust test accuracy of EFNT+AT increases 10\%-18\% when defending adversarial examples generated by PGD-10 and PGD-20.
For the comprehensive experiments in Table \ref{table:beta5} and Table \ref{table:beta10}, it is easy to verify that our proposed model outperforms all other methods regardless of the value of $\alpha$.

To sum up, client deep models trained by EFAT with $\alpha=5\%$ have higher robust test accuracy but lower standard test accuracy. By increasing $\alpha $ to 10\%, client deep models have slightly increased on both standard test accuracy and robust test accuracy.

\section{Conclusion}
% This paper proposed the Ensemble Adversarial Federated Training method to improve the robustness of models against black-box attacks with Non-IID training data. Our approach improves the federated model robustness through adversarial training by enhancing the diversity of adversarial examples through expanding training data with perturbations generated from other participated clients.
% Experiment results show that compared with the intuitive federated adversarial training method and the other two variants of EFAT, our model makes a significant improvement on robustness and accuracy on both Digit-Five and CIFAR10 in IID and Non-IID settings. We believe the further exploration of this direction will lead to more findings on the robustness of federated learning.

In this paper we present a novel ensemble federated adversarial training method, termed as EFAT, to improve the robustness of models against black-box attacks in federated learning. The proposed method enhances the diversity of adversarial examples through expanding training data with perturbations generated from other participating clients. 

Experiment results on both Digit-Five and CIFAR10 in IID and Non-IID settings show that our method significantly improves the robustness and accuracy contrasted with the intuitive federated adversarial training method and the other two variants of EFAT. 

\clearpage

% \bibliography{references}
% \cite{li2020federated}
% \cite{mcmahan2017communication}
% \cite{bonawitz2017practical}
% \cite{bagdasaryan2020backdoor}
% \cite{bhagoji2019analyzing}
% \cite{szegedy2013intriguing}
% \cite{goodfellow2014explaining}
% \cite{koh2017understanding}
% \cite{papernot2017practical}
% \cite{tramer2018ensemble}
% \cite{madry2018towards}
% \cite{carlini2017towards}
% \cite{suciu2018does}
% \cite{dwork2008differential}
% \cite{mcmahan2017learning}
% \cite{augenstein2019generative}
% \cite{athalye2018obfuscated}
% \cite{xie2019feature}
% \cite{shafahi2019adversarial}
% \cite{kairouz2019advances}
% \cite{kurakin2016adversarial}

\bibliographystyle{named}
\bibliography{ijcai21}  %%% Remove comment to use the external .bib file (using bibtex).

\end{document}